# Control Strategies for Mobile Robot With Obstacle Avoidance

**M. Zohaib[1], M. Pasha[1], R. A. Riaz[1], N. Javaid[1], M. Ilahi[1], R. D. Khan[2]**

[1]COMSATS Institute of Information Technology, Islamabad, Pakistan.
[2]COMSATS Institute of Information Technology, Wah Cant, Pakistan.

**ABSTRACT**

Obstacle avoidance is an important task in the field of robotics, since the goal of autonomous robot is to reach the destination without collision. Several algorithms have been proposed for obstacle avoidance, having drawbacks and benefits. In this survey paper, we mainly discussed different algorithms for robot navigation with obstacle avoidance. We also compared all provided algorithms and mentioned their characteristics; advantages and disadvantages, so that we can select final efficient algorithm by fusing discussed algorithms. Comparison table is provided for justifying the area of interest

## I. INTRODUCTION

Obstacle avoidance is the back bone of autonomous control as it makes robot able to reach to destination without collision. Path planning is involved to generate the shortest path from source to destination on the basis of sensorial information of environment. Many obstacle avoidance algorithms are proposed, some of them are discussed in this paper. Bug algorithms are the earliest methods [1]. They are easy to tune but more time consuming. They are not goal oriented algorithms, as they follow the edge without considering the goal. Same as, Artificial Potential is also a easy technique for obstacle avoidance but they get stuck in local minima [1][2]. Vector Field Histogram (VFH) is used by [2][8], that is an improved algorithm. It selects a shorter path than bug algorithms but it takes more time to manipulate. Follow the gap (FGM) method is a novel algorithm that is proposed in 2012 but it also unable to avoid U-shaped obstacle [1]. New Hybrid Navigation algorithm (NHNA) is a complete algorithm, which proves convergence but it is unable to apply in an unknown environment as it requires prior information of environment [3]. Same as "NHNA", a Hybrid Navigation Algorithm (HNA) with roaming trails is an obstacle avoidance algorithm for partially known environment [4]. It used APF in its reactive layer so it can also get stuck in local minima. It is also a time consuming algorithm as robot may stop in front of obstacle until it moves. The characteristics of different algorithms are compared in table1. The main algorithms that we have studied are discussed in coming sections (Latest Journal and research papers are preferred to study).

## II. ARTIFICIAL POTENTIAL FIELD METHOD

This algorithm is based on the principle of Potential field in which robot and obstacles are act as a positive charge where as goal act as a negative charge. Thus, obstacles repel robot by generating repulsive force and goal attracts robot due to opposite change. Final force on robot is the vector sum of all repulsive and attractive force. However the magnitude of force is described by the distance, i.e. the obstacle near to robot will affect more similarly when the robot is at a far distance from goal its speed will be high and it will become slow as it comes close to goal. As mention in [2] attractive force is –ve gradient of attractive potential.

$$F_{attr} = -\nabla U_{attr} = -K_{attr}\,(q - q_{goal})$$

Where $q - q_{goal}$ is Euclidean distance from current position to goal and $K_{attr}$ is scaling factor.

***Corresponding Author:** Nadeem Javaid, COMSATS Institute of IT, Islamabad, www.njavaid.com.

Repulsive force can be calculated by adding a repulsive effect on robot by the obstacles. This can be done by calculating the distance of obstacles from robot and their direction (angle). The obstacle near to robot has high repulsive force. The formula that [3] described is,

$$U_{rep} = \sum_{i=1}^{n} U_{rep\ i}\ (q)$$

The –ve gradient of $U_{rep}$ is a repulsive force. So,

$$F_{rep} = -\ U_{rep\ i}\ \ (q)$$

APF is a goal oriented algorithm and selects shorter path from source to destination, however it has a local minima problem. Symmetric and U-shaped obstacles are the dead end scenarios for APF as illustrated in Fig.1. Symmetric obstacles are shown in Fig.1a, in which attractive force of goal is equal and opposite to the sum of repulsive forces by obstacles. So the final heading force becomes zero and robot stops its motion, this is the case of local minima. Another crucial scenario is shown in Fig.1b, which also cases the local minima and APF fails to avoid it.

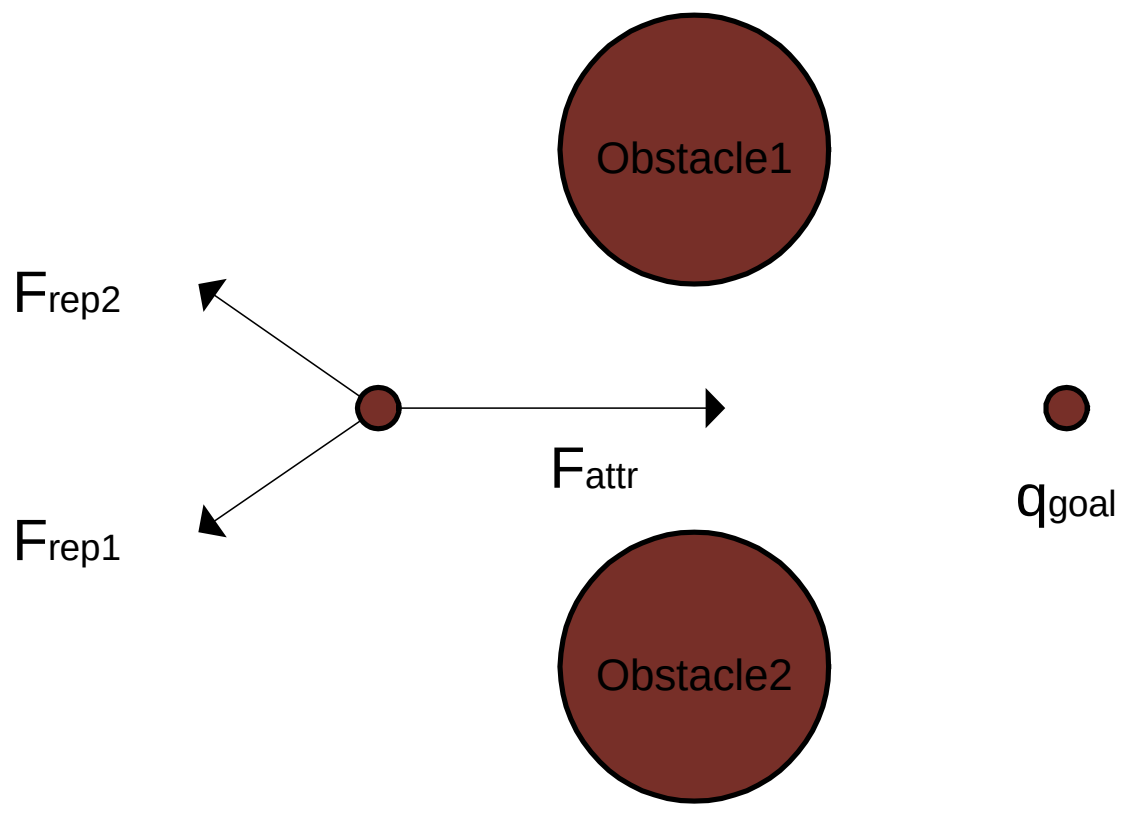

Fig.1. Dead end scenario of Artificial Potential Field method (symmetric obstacles)

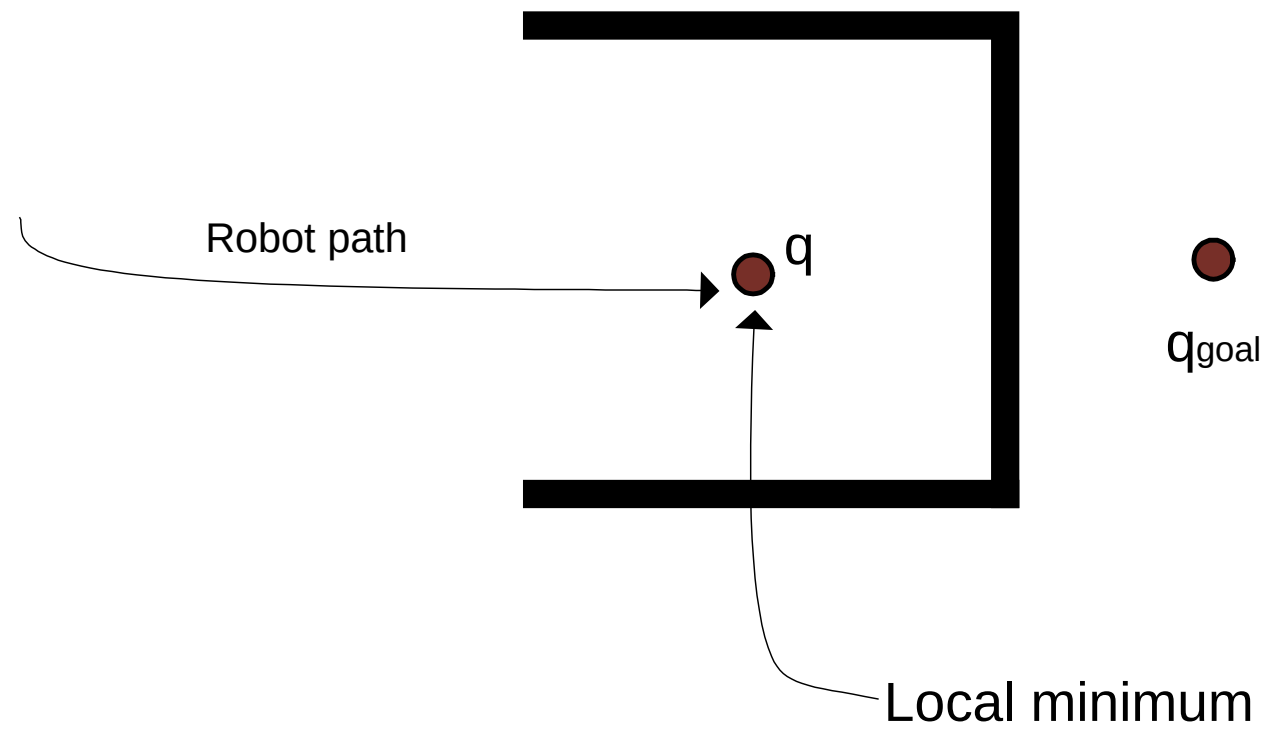

Fig.2. Dead end scenario of Artificial Potential Field method (U-shaped obstacle)

## III. VECTOR FIELD HISTOGRAM

Vector field Histogram is a three stage method of obstacle avoidance. In first stage 2D histogram is generated around the robot that represents the obstacles. 2D histogram is updated with new coming percepts from sensors. In the second step, this 2D histogram is converted to 1D histogram and then polar histogram. Finally in a third step, the algorithm selects the most suitable sector with low polar obstacle density, and calculates the steering angle and velocity in that direction. The Fig.2 is taken by the work of [8] which illustrates the 2D histogram grid. Conversion from 2D to 1D histogram is depicted in Fig. 3a and Fig. 3b is the representation of 1D polar histogram.

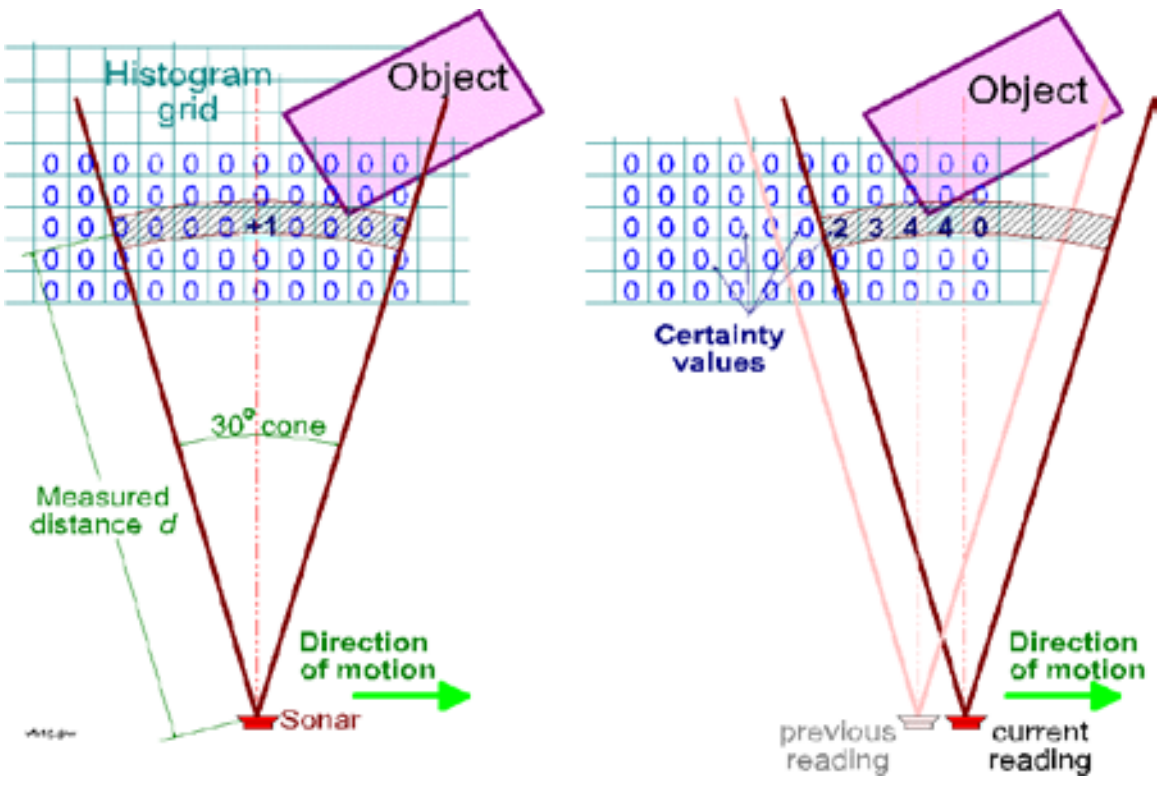

Fig. 3. Construction of 2D histogram grid map

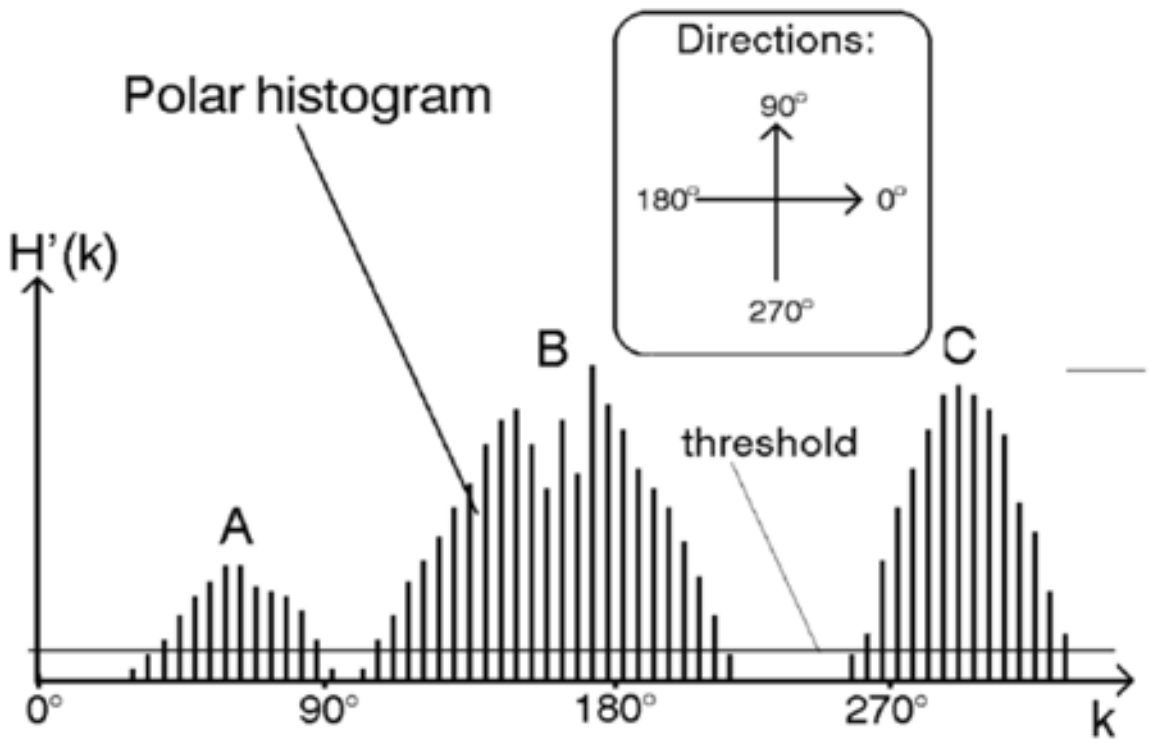

Fig. 4. Representation of 1D and polar histogram (1D Histogram)

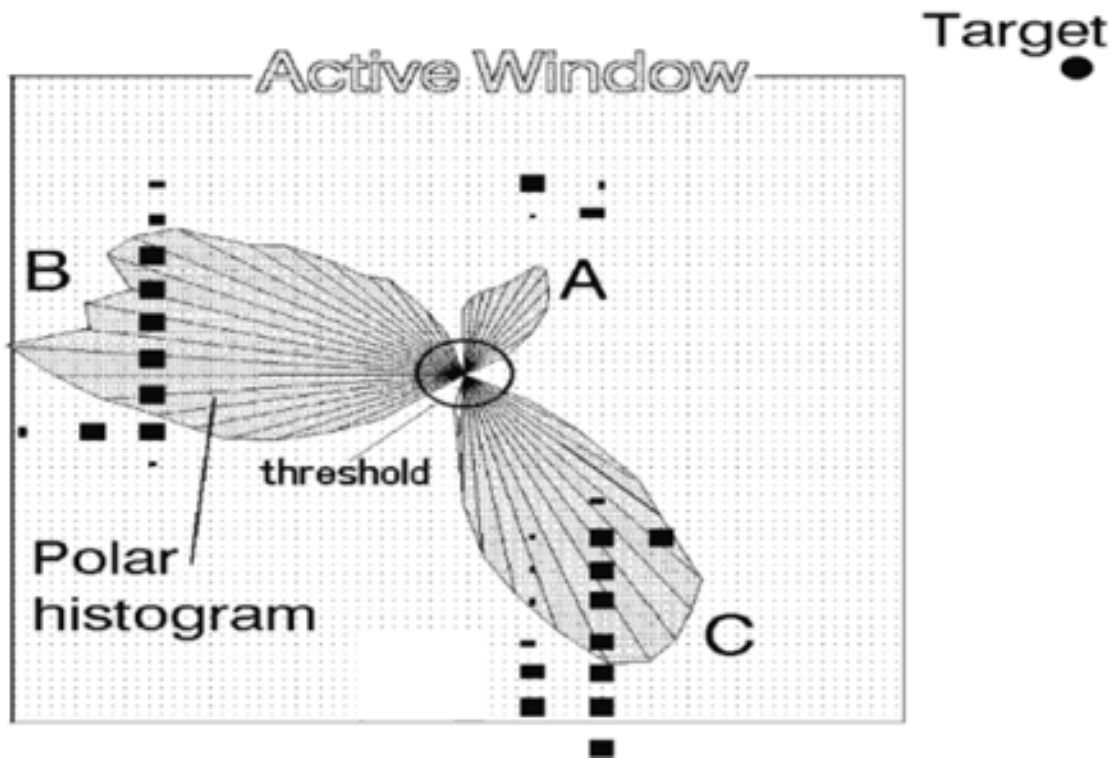

Fig. 5. Representation of 1D and polar histogram (1D Polar histogram)

## IV. BUG ALGORITHM

One of the earliest algorithms is Bug algorithm, which plans direct path from source to destination until it faces an obstacle. Algorithm is sub divided into three main versions on the basis of their behaviour of obstacle avoidance is mentioned below;

- **VERSION 1: BUG-1 ALGORITHM**

In this algorithm when robot detects an obstacle it start moving around it until reaches to starting point from where it has started. During its movement around an obstacle, it calculates a leaving point with minimum distance to destination and generates new path from calculated leaving point to destination. After its one complete circle, it restarts its motion around obstacles until reaches to leaving point and starts moving on new generated path to reach the destination. Fig. 4a is the simulation results of [7] that shows the trajectory of robot under bug1 algorithm.

- **VERSION 2: BUG-2 ALGORITHM**

Bug-2 algorithm generates slope from an initial position to destination and robot starts following it until it interrupted by obstacle. When it interrupted, it follows the edge of obstacle and calculates new slop from every new position until the new slop becomes equal to the original slope. After reaching on point having same slope as previous, it starts moving to destination by following pervious generated path. The scenario is illustrated in Fig. 4b [7].

- **VERSION 3: DIST-BUG ALGORITHM**

This algorithm is based on distance, in which robot moves from source to destination on path having minimum distance. When robot faces an obstacle in path, it starts following the edge of obstacle simultaneously; it calculates the distance of destination from each point. The point with the minimum distance is known as leaving point. When it finds the leaving point during its motion around an obstacle, it generates a new path and starts following it until reaches to destination as shown in Fig. 4c [7].

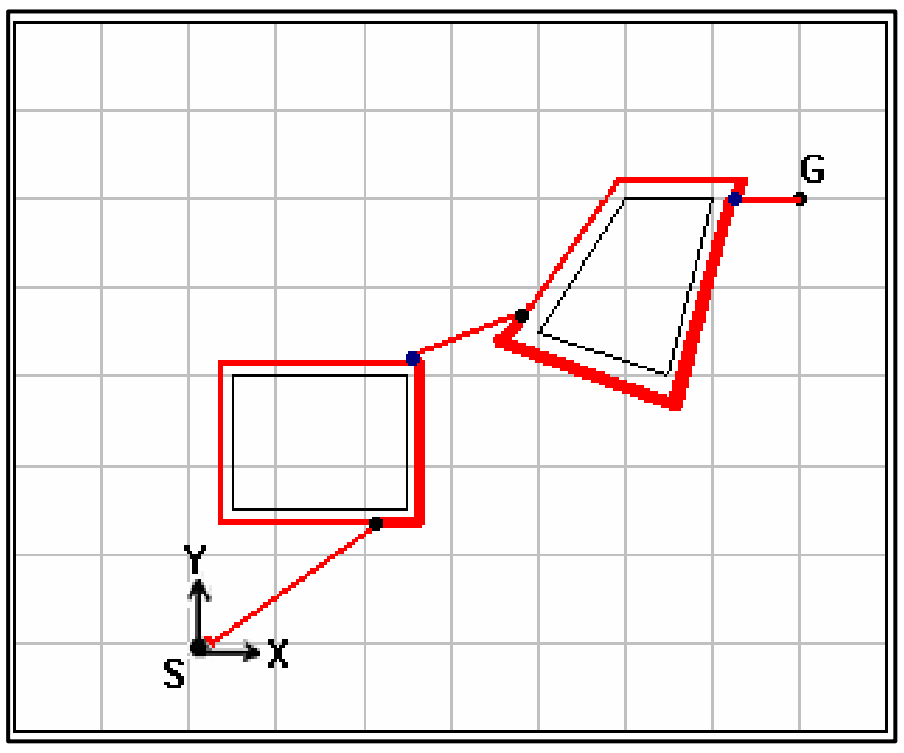

Fig.6. Obstacle avoidance with Bug algorithms (Trajectory of Bug-1 algorithm)

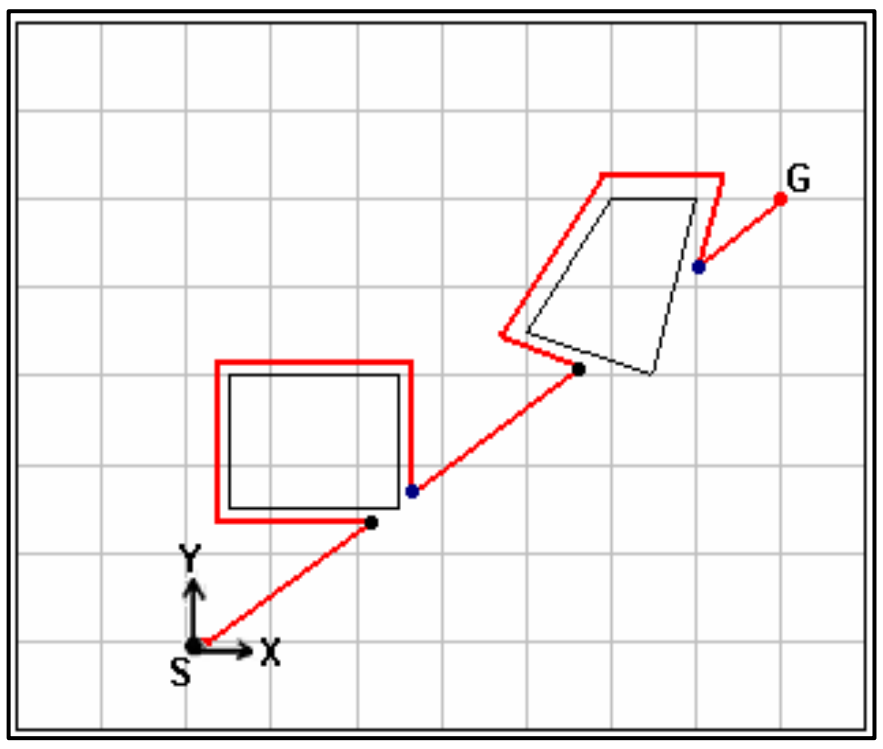

Fig.7. Obstacle avoidance with Bug algorithms (Trajectory of Bug-2 algorithm)

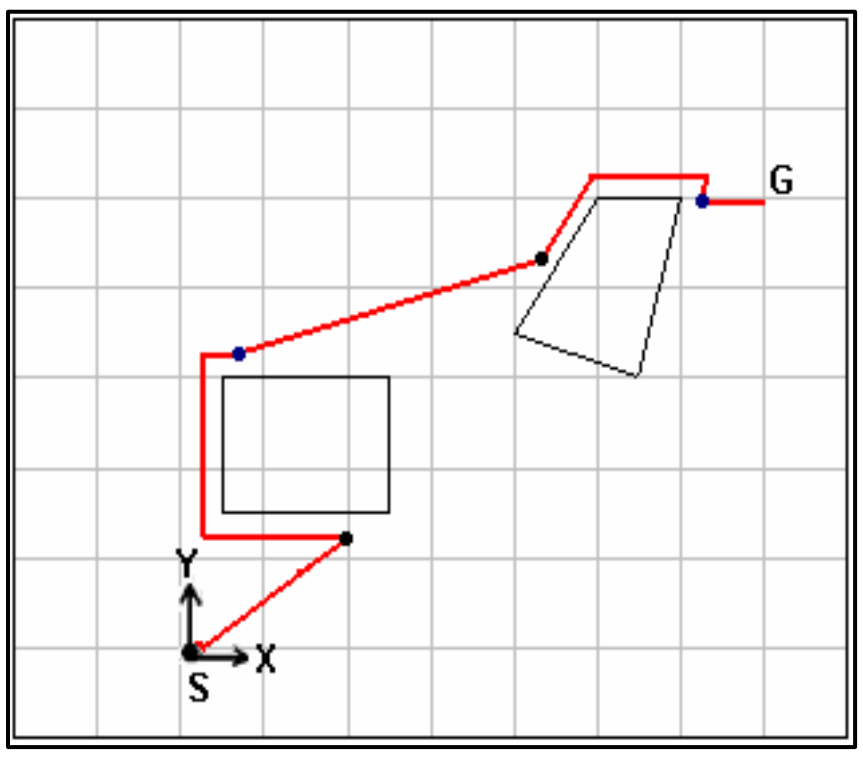

Fig.7. Obstacle avoidance with Bug algorithms (Trajectory of Dist-Bug algorithm)

## V. FOLLOW THE GAP METHOD (FGM)

Follow the gap method avoids obstacles by finding the gap between them. It calculates the gap angle. It has a threshold gap, the minimum gap between obstacles from which robot can move. If the measured gap is greater than the threshold gap then robot will follow calculated gap angle. Obstacle avoiding using “FGM” is done in three main steps.

- **STEP-1: CALCULATING THE GAP ARRAY AND FINDING THE MAX. GAP**

In step 1, When robot face obstacles it calculates the distance of obstacle from robot and stores these distance in distance array. After finding the distances of all obstacles, gap array is generated, which includes the gap between obstacles. Gap array is being traversed to find a maximum gap between obstacles. If more than one Maximum gap exists with the same value, then first gap will be selects as a maximum gap. The method used by author, to generate gap array is shown in Fig. 5a [1]. The pink lines are indicating the nonholonomic constraints of robot where as doted green lines are the field of view of robot. $d_{nhol_l}$ and $d_{nhol_r}$ are the distances of obstacles from left and right nonholonomic constraints lines and $d_{fov_l}$ and $d_{fov_r}$ are the distances of obstacles from left and right field of view lines respectively. $\emptyset_{nhol_l}$ and $\emptyset_{nhol_l}$ are the angles of left and right nonholonomic constraint lines and $\emptyset_{fov_l}$ and $\emptyset_{fov_r}$ are the angles of left and right field of view lines of robot. The distance with less value is stored and avoided first.

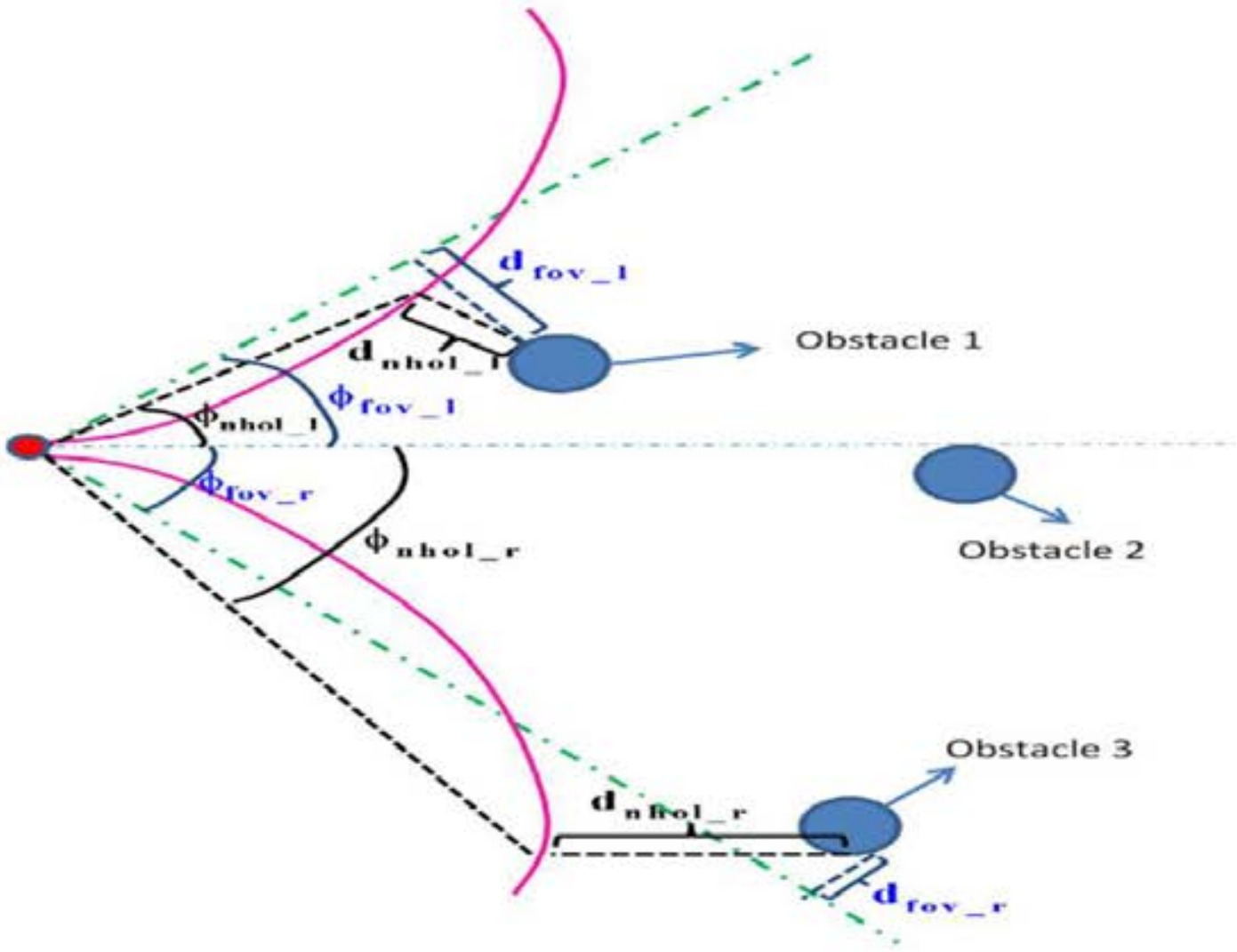

Fig.8. Representation of Gap border parameter and center gap angle (Gap border parameters)

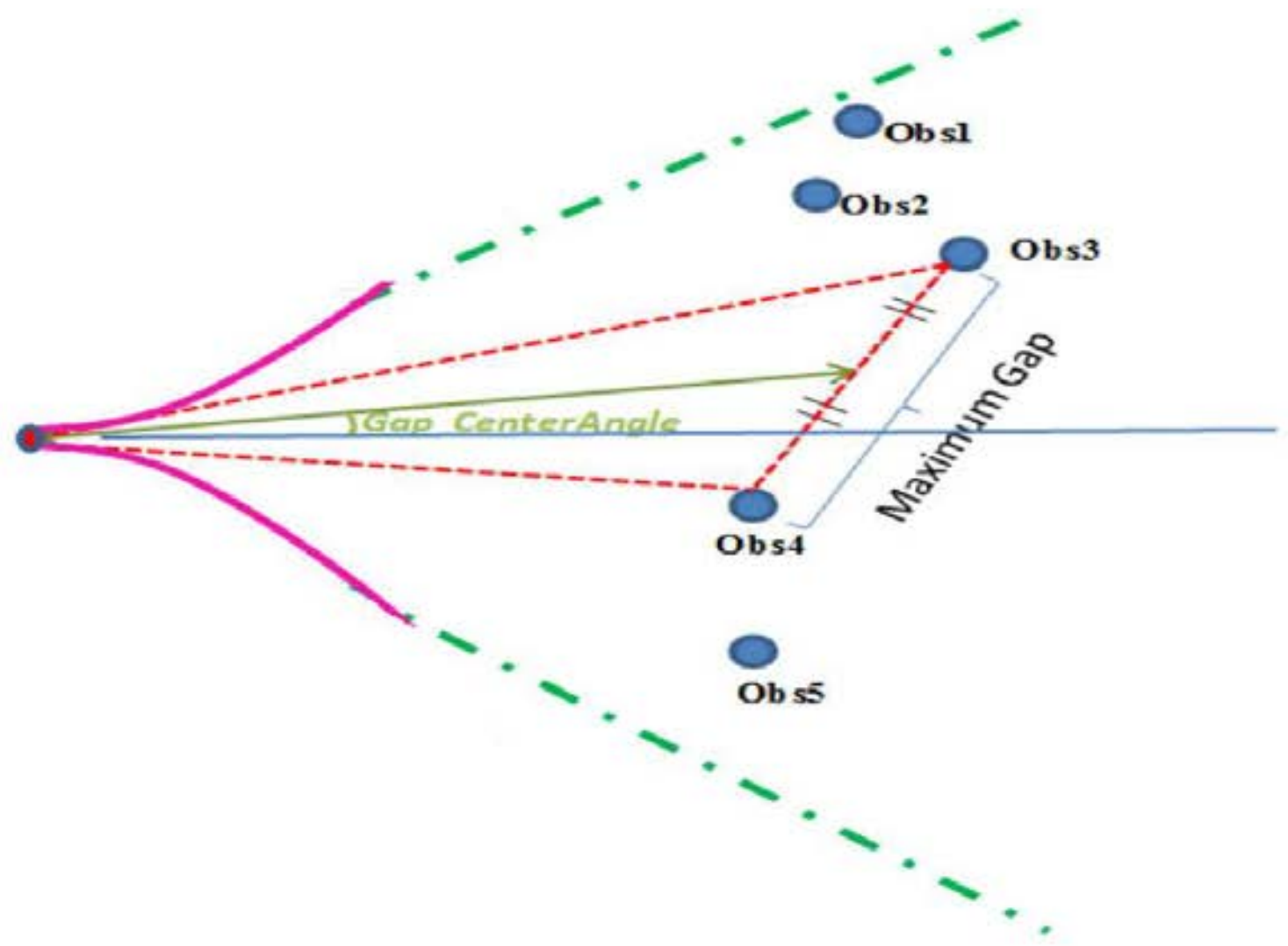

Fig.9. Representation of Gap border parameter and center gap angle (Center gap angle)

- **STEP-2 : CALCULATION OF GAP CENTER ANGLE**

The second step of FGM is to calculate center angle of the maximum gap, which ensures the safe trajectory from the center of obstacles. This is an angle of vector having tail at robot's current position and head on the center point of maximum gap. Gap center angle can be calculated by using Apollonius theorem and law of cosine as being done by [1]. Final equation of $\emptyset_{gap_c}$ is shown below;

$$\emptyset_{gap_c} = \boldsymbol{arccos}\left(\frac{\boldsymbol{d1+d2\,cos(\emptyset_1+\emptyset_2)}}{\sqrt{\boldsymbol{d_1^2+d_2^2+2d_1d_2\,cos(\emptyset_1+\emptyset_2)}}}\right) - \emptyset_{\mathbf{1}} \quad (111)$$

Where $\boldsymbol{d_1}$ and $\boldsymbol{d_2}$ are the distances of obstacle 1 and 2 from robot respectively. $\emptyset_{\mathbf{1}}$ ***and*** $\emptyset_{\mathbf{2}}$ are angles of obstacle 1 and 2 respectively. $\emptyset_{gap_c}$ is the final calculated gap center angle. Fig. 5b illustrates the gap center angle.

- **Step-3 : Calculation the final heading Angle**

The last stage of "FGM" is to calculate the final heading angle. This can be achieved by combining the gap center angle with the goal angle. The combining structure is distance of obstacles and weight dependent, i.e. the obstacle nearer to robot has more weight. In case, when obstacle is at very short distance to robot then robot must move to gap angle rather than goal angle. It is due to fact that obstacle avoidance is the main task of path planning. Formula to calculate the final angle is given bellow:

$$\emptyset_{final} = \frac{\frac{\alpha}{d_{min}}\emptyset_{gap_c} + \beta\,\emptyset_{goal}}{\frac{\alpha}{d_{min}} + \beta} \quad (222)$$

Where, $\boldsymbol{d_{min}} = \min_{i=1:n}(\boldsymbol{d_n})$, $\emptyset_{gap_c}$ and $\emptyset_{goal}$ are calculated gap and goal angle, $\boldsymbol{\alpha}$ and $\boldsymbol{\beta}$ areweight coefficients of gap and goal angle respectively. (For simplicity, $\boldsymbol{\beta}$ can be consider 1).

In short "FGM" when robot encounters the obstacle, it starts finding the gap between obstacles and save these calculated gap vales in an array. Algorithm finds the maximum gap from calculated gap array. If the maximum gap is greater than the threshold value then it calculates the gap angle, while simultaneously it considers goal angle. Finally gap angle is added into goal angle with their weight coefficients to find final angle to avoid obstacles. After these entire calculations robot starts moving along final calculated angle in order to avoid an obstacle.

## VI. NEW HYBRID NAVIGATION ALGORITHM (NHNA)

"New Hybrid navigation" algorithm based on two layers, deliberative layer and reactive layer. Both layers are independent to each other. Deliberative layer planed a reference path on the basis of stored prior information. Reactive layer is an independently steers robot on the path planed by the deliberative layer.

Hybrid algorithm required prior information of environment, which is stored in the form of binary grid map. In map, states of every grid are either free of occupied that depends on obstacles around i.e. free for no obstacle and occupied for obstacle. Unknown information is also taken as a free. In deliberative layer, A* search algorithm is used to generate a reference path. Reference path is temporary and not necessary to follow through out motion, it can be changed by the reactive layer. Fig. 6 is the results of [3] which shows the planned and shortest paths generated by A* search algorithm.

Reactive layer takes reference path from deliberative layer and controls the motion of robot. It also receives the percepts of sensors and take decision to avoid an obstacle if found. For the purpose of obstacle avoidance, this layer uses D-H bug algorithm (Distance Histogram bug). This is a version of bug-2 algorithm which is improved by [2], which allows robot to rotate freely at angle less than $90^{\circ}$ to avoid an obstacle. If the rotation of $90^{\circ}$ or greater is required to avoid an obstacle; it acts as bug-2 algorithm and starts moving to destination when path is clear from obstacles. Fig. 7a shows the robot trajectory with Dist-Bug algorithm where as Fig. 7b illustrates the robot's behavior with D-H bug algorithm [3].

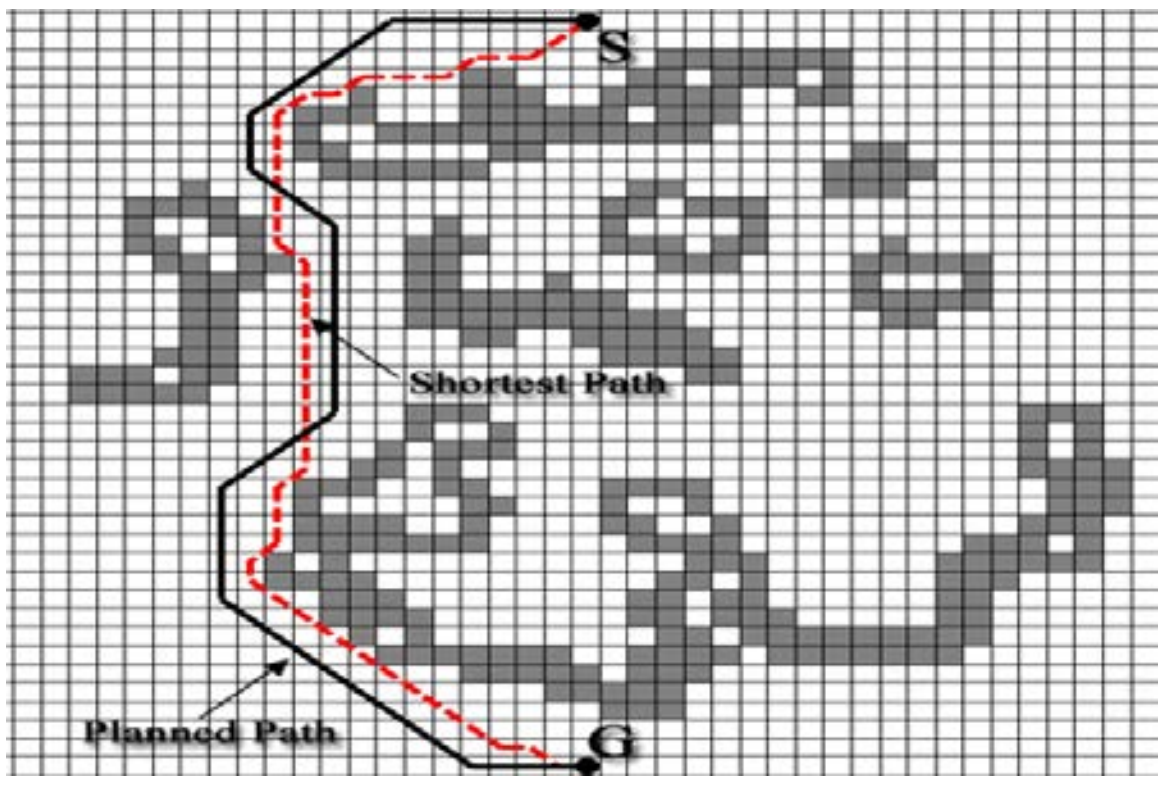

Fig.10. Grid Map and Trajectory of robot with (NHNA)

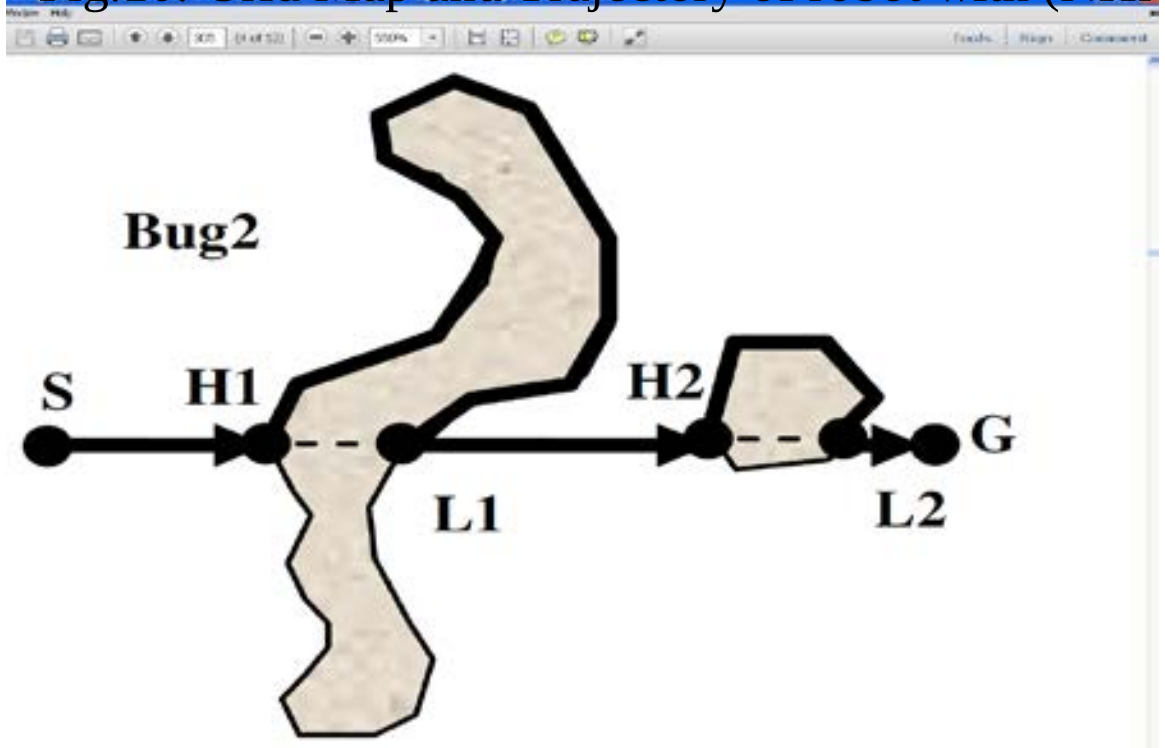

Fig. 11: strategy of obstacle avoidance (Trajectory of Dist-bug algorithm)

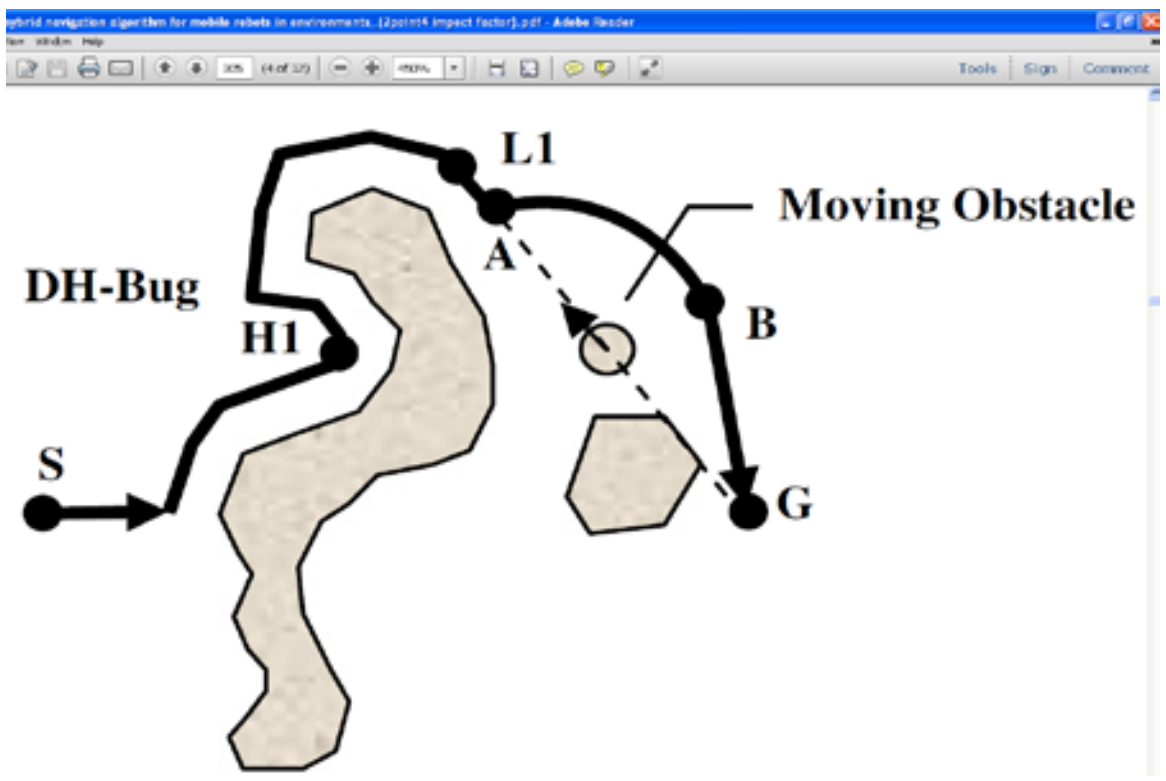

Fig. 11: strategy of obstacle avoidance (Trajectory of robot with Distance Histogram (D-H) bug algorithm)

Reactive layer can change the path on the basis of current percept. Sensors provide current percepts to reactive layer as well as it updates the prior knowledge. In case on conflict between layers, the result of reactive layer is taken into an account. It is due to the present and updated nature of the results of reactive layer and hence incomplete knowledge of deliberative layer may contain errors.

## VII. ZYBRID NAVIGATION ALGORITHM WITH ROAMING TRAILS (HNA)

The Hybrid Navigation algorithm with roaming trails is related to new NHNA. The main difference is that it used APF instead of D-H BUG in reactive layer. NHNA has not described any limit for robot to deviate from reference path but HNA used the concept of roaming trails for the same purpose. Fig. 8a shows the roaming trails with prior map and Fig. 8b illustrates the safe trajectory of robot in roaming trails [4].

According to the work of [4], other Hybrid algorithms may get stuck into cul-de-sac scenario as shown by the table in Fig. 8b. However it may stops in front of obstacle until it moves.

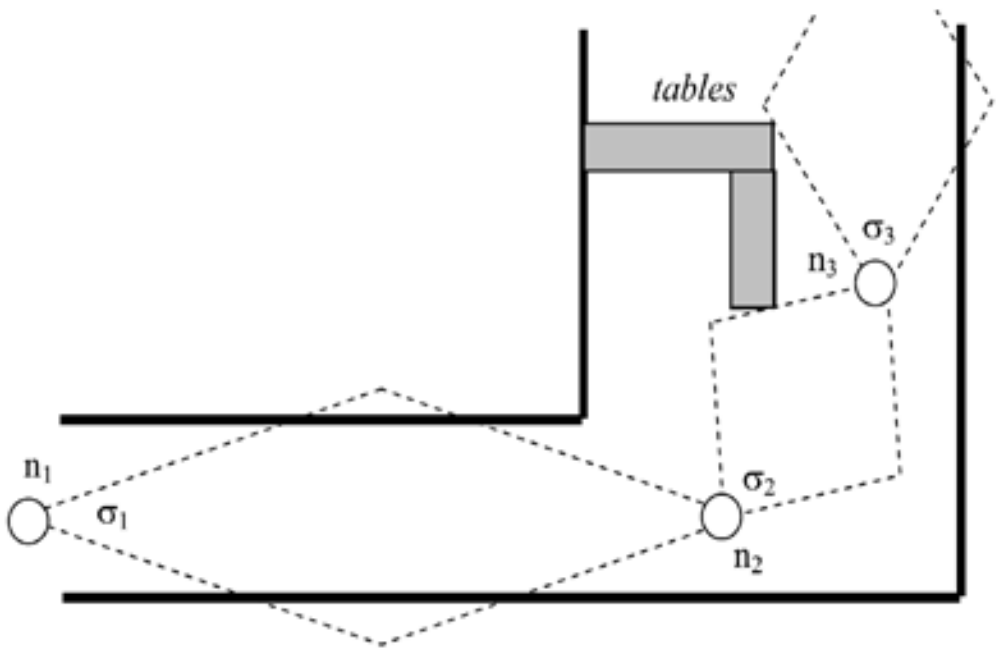

Fig.12. Trajectory of robot with Roaming Trails (HNA) (Priori map with Roaming Trails)

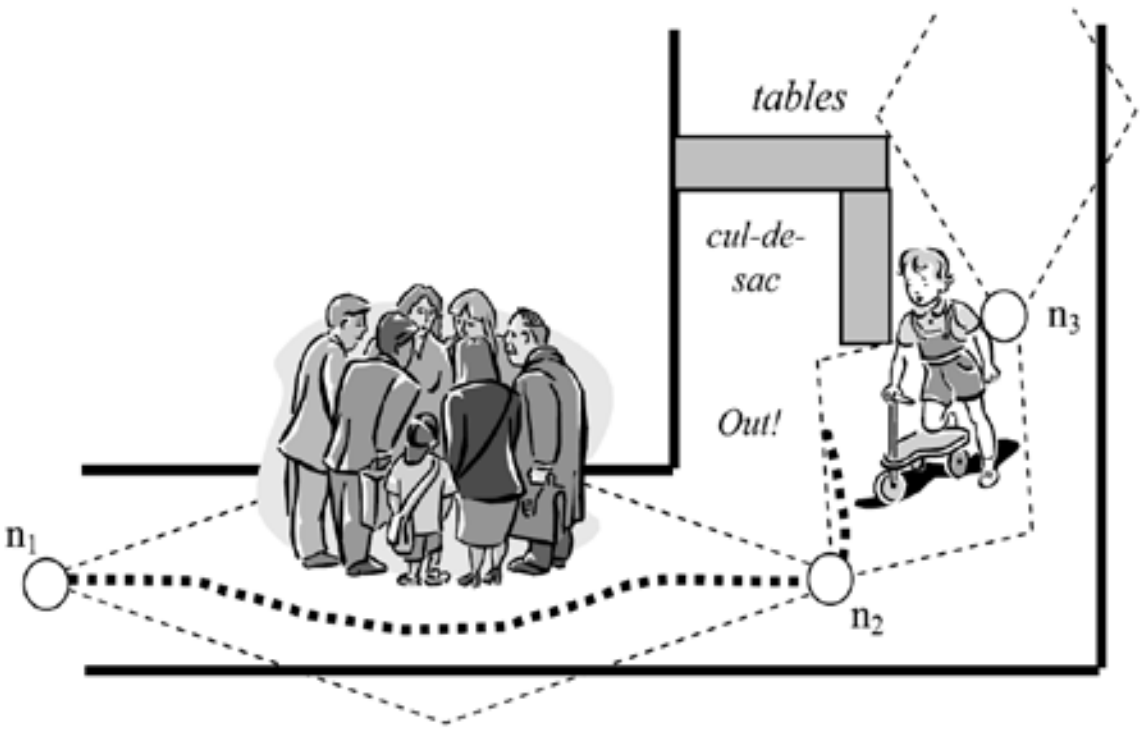

Fig.13.Trajectory of robot with Roaming Trails (HNA) (Trajectory of the robot (dotted line))

## VIII. COMPARISON BETWEEN ALGORITHMS

The above mentioned algorithms on obstacle avoidance are different with each other in some aspects. The main characteristics are compared and depicted in table shown in table-1.
“Dist-Bug algorithm” is efficient then “Potential Field Method” because it has no local minimum problem. It covers short distance as compared to previous versions but it may take robot away from the goal because it is not a goal oriented while following the wall. Where as Potential Field Method is easy to tune, but it is not preferred since it get stack into local minimum error. Vector Field Histogram is time, space consuming algorithm as in 1st step, it generates 2D histogram and then converted it into 1D histogram for further calculation.
“Vision based sensor” method is best for upper level control i.e. with the use of beagle board, FPGA or any high processor because it requires dedicated application, high memory and calculations. Since we are working with Micro controller which is unable to interface camera, and cannot run any software relating to image processing.

| **Algorithm** | **Implementation** | | **Performance** | | | **Remarks** |
|---|---|---|---|---|---|---|
| | **Hardware Required** | **Parameters Required** | **Efficiency** | **Convergence** | **Time Complexity** | |
| Bug Algorithm [2][7] | Distance sensors (IR, sonar) Microcontroller | Current and destination position | Low, may take robot away from destination | Yes, but take more time to achieve goal | Always move in one direction to avoid obstacle which increases the time complexity | No local minima occur Select longest path |
| Potential Field Method (VFF) [2][10] | Distance sensors(IR, sonar) Microcontroller | Target and obstacle distance | Low, calculation are not accurate, constraints are not taken into account | No, (in case U-shaped and symmetrical obstacles) | Less time required as it selects shorter path | Local minima can occur |

*Zohaib et al.*,2013

| | | | | | | |
|---|---|---|---|---|---|---|
| Vector Field Histogram [8] | Sonar sensor Processor, high memory | Obstacle distance | Low, calculation may accurate but consumes more resources like memory, processor and power | No, (in case U-shaped and symmetrical obstacles) | Required more time to generate a 2D grid and conversion from 2D to 1D polar histogram | Difficult for Microcontroller as high computations are required |
| Vision sensor based method [11] | Camera, sonar sensor, Processor, beagle board, laptop | Obstacles position, angle and distance | High, calculations are real and accurate (depends on equipment) | May or may not (depend on Nature of algorithm) | Depends on the resolution of camera and application used, mostly take more time for calculations | Not best for mini vehicle with micro controller, It requires laptop or Processor and specific application like MATLAB. |
| Follow the Gape Method [1] | Ultrasonic and lidar Sensors, camera optical velocity sensor, NIPXI-811108RT processor, PXi-7954R FPGA | Obstacle distance and angle | High, Easy to tune, always select shortest path, able to avoid symmetric obstacles | No, (In dead end scenario like U-shaped obstacle) | Less time consuming as decision are made on the basis of currents percepts, | Fails for U shaped obstacle |
| Hybrid Navigation Algorithm With Roaming trails [4] | Laser and sonar sensors, Pioneer 3-AT robots | Prior information of information | Medium, generates shortest path but no limit to deviate from path | Yes, but in some scenarios, robot may stop in front of obstacle | Consume more time generate reference path (A* search is required) Minutes or seconds | Requires High calculation Consume more time In seconds and minutes |
| New Hybrid Navigation Algorithm (NHNA) [3] | Laser sensor, Micro processor | Prior information of information | Medium and efficient then Hybrid, use DH-Bug algorithm | Yes, mostly converges except some scenarios like cul-de-sac | Consume more time as A* search is required to generate path | High calculation Consume more time In seconds |

## IX. CONCLUSION

Collision free algorithm is a requirement of autonomous vehicle, since it provides the safe trajectory and proves the convergence. Some of the main algorithms that can use for obstacle avoidance are discussed in this paper. Dist-bug algorithm is an efficient algorithm in Bug series but it still takes more time to reach to destination. It is not a goal oriented; it may take robot for away from goal position. It can be improve by applying some condition as, if path is free towards goal, stops edge detecting and regenerate new path to move forward. From the above table, we conclude that "Follow the gap" method is better algorithm than others since it takes less time to reach the destination and does not require any dedicated software or extra memory. It has an important problem (due to its local characteristics), as it is unable to avoid U and H-shaped obstacles. So, there is a need of an algorithm which cannot get stuck into local minima and can able to tackle obstacles of U and H shaped. This can be achieved by fusing discussed algorithms with some upper level intelligence. This is our future task to design an algorithm that can avoid U and H-shaped obstacles and has no local minima issue.